\documentclass[11pt,a4paper]{article}
\usepackage[hyperref]{acl2019}
\usepackage{times}
\usepackage{latexsym}
\usepackage{amsmath}
\usepackage{url}
\usepackage{fancyvrb}
\usepackage{booktabs} 

\aclfinalcopy 
\def\aclpaperid{2285} 

\usepackage{soul}
\usepackage{graphicx}
\usepackage{subfigure}
\usepackage{amsfonts}
\usepackage{multirow,multicol}
\usepackage{amsmath} 

\title{Unified Semantic Parsing with Weak Supervision}

\author{Priyanka Agrawal\textsuperscript{*}, Parag Jain, Ayushi Dalmia,\\ \textbf{Abhishek Bansal, Ashish Mittal, Karthik Sankaranarayanan} \\
  IBM Research AI \\
\textsuperscript{*}\small{\texttt{pagrawal.ml@gmail.com}}\\
 \small{\texttt{\{pajain34, adalmi08, abbansal, arakeshk, kartsank\}@in.ibm.com}} \\
  }

\date{}

\begin{document}
\maketitle
\begin{abstract}
Semantic parsing over multiple knowledge bases enables a parser to exploit structural similarities of programs across the multiple domains. However, the fundamental challenge lies in obtaining high-quality annotations of (utterance, program) pairs across various domains needed for training such models. To overcome this, 
we propose a novel framework to build a unified multi-domain enabled semantic parser trained only with weak supervision (denotations).  Weakly supervised training is particularly arduous as the program search space grows exponentially in a multi-domain setting.
To solve this, we incorporate a multi-policy distillation mechanism in which we first train domain-specific semantic parsers (teachers) using weak supervision in the absence of the ground truth programs, followed by training a single unified parser (student) from the domain specific policies obtained from these teachers.  
  The resultant semantic parser is not only compact but also generalizes better, and generates more accurate programs. It further does not require the user to provide a domain label while querying. On the standard \textsc{Overnight} dataset (containing multiple domains), we demonstrate that the proposed model improves performance by 20\% in terms of denotation accuracy in comparison to baseline techniques.

\end{abstract}

\begin{figure}
\centering
\includegraphics[scale=0.38]{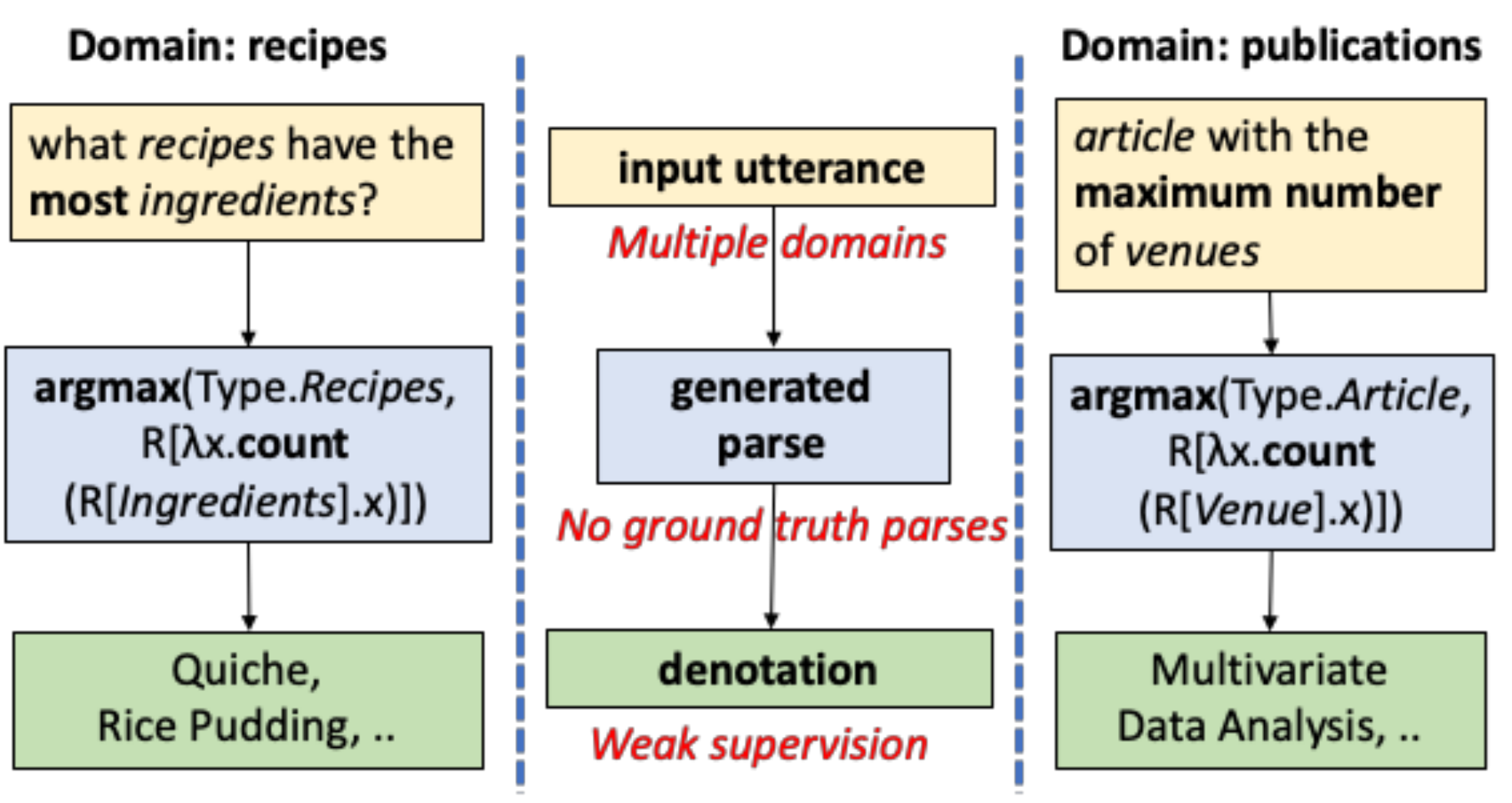}
\caption{Examples for natural language utterances with linguistic variations in two different domains
that share structural regularity (Source:  \textsc{Overnight} dataset). Note that in this setup, we do not use ground truth parses for training the semantic parser.}
\label{fig:example}
\end{figure}

\section{Introduction}
Semantic parsing is the task of converting natural language utterances into machine executable programs such as SQL, lambda logical form~\cite{liang2013lambda}. This has been a classical area of research in natural language processing (NLP) with earlier works primarily utilizing rule based approaches~\cite{Woods1973} or grammar based approaches~\cite{Lafferty:2001,Kwiatkowski2011, Zettlemoyer:2005,Zettlemoyer:2007}.
Recently, there has been a surge in neural encoder-decoder techniques which are trained with input utterances and corresponding annotated output programs~\cite{DongP16-1004,P16-1002}. However, the performance of these strongly supervised methods is restricted by the size and the diversity of training data i.e. 
natural language utterances and their corresponding annotated logical forms. This has motivated the work on applying
weak supervision based approaches \cite{clarke2010driving, LiangNSM, Neelakantan2016LearningAN, P18-1071}, which use \emph{denotations} i.e. the final answers obtained upon executing a program on the knowledge base and use \textsc{REINFORCE} \cite{williams1992simple, NIPS2016_6547}, to guide the network to learn its semantic parsing policy (see Figure~\ref{fig:subfigure1_framework}). Another line of work~\cite{goldman2018weakly, Cheng2018WeaklysupervisedNS} is aimed towards improving the efficiency of weakly supervised parsers by applying a two-stage approach of first learning to generate program templates followed by exact program generation. It is important to note that this entire body of work on weakly supervised semantic parsing has been restricted to building a parser over a single domain only (i.e. single dataset).

Moving beyond single-domain to multiple domains,  \newcite{Herzig2017NeuralSP} proposed semantic parsing networks 
trained by combining the datasets corresponding to multiple domains into a single pool. Consider the example in Figure \ref{fig:example} illustrating utterances  from  two domains, 
\textsc{recipes} and \textsc{publications},
of the \textsc{Overnight} dataset. The utterances have linguistic variations \textit{most} and \textit{maximum number} corresponding to the shared program token \textit{argmax}.  This work shows that leveraging such structural similarities in language by combining these different domains leads to improved performance.  However, as with many single-domain techniques, this work also requires strong supervision in the form of program annotations corresponding to the utterances.
Obtaining such high quality annotations across multiple domains is challenging, thereby making it expensive to scale to newer domains. 

To overcome these limitations, in this work, we focus on the problem of developing a semantic parser for multiple domains in the weak supervision setting using denotations. Note that, this combined multiple domain task clearly entails a large set of answers and complex search space in comparison to the individual domain tasks.   
Therefore, the existing multi-domain semantic parsing models \cite{Herzig2017NeuralSP} fail when trained under weak supervision setting.
See Section \ref{ssec:result} for a detailed analysis.

To address this challenge, we propose a {multi-}policy distillation framework for multi-domain semantic parsing. This framework splits the training in the following two stages: 1) Learn domain experts (teacher) policy using weak supervision for each domain. This allows the individual models to focus on learning the semantic parsing policy for corresponding single domains; 
2) Train a unified compressed semantic parser (student) using distillation from these expert policies.
This enables the unified student to gain supervision from the above trained expert policies and thus, learn the shared semantic parsing policy for all the domains. 
This two-stage framework is inspired from policy distillation \cite{rusu-distillation-2015} which transfers policy of a reinforcement learning (RL) agent to train a student network that is more compact and efficient. In our case, weakly supervised domain teachers serve as RL agents.
For inference, only the compressed student model is used which takes as input the user utterance from any domain and outputs the corresponding parse program. It is important to note that, the domain identifier input is not required by our model.  The generated program is then executed over the corresponding KB to retrieve denotations that are provided as responses to the user. 


To the best of our knowledge, we are the first to propose a unified multiple-domain parsing framework which does not assume the availability of ground truth programs. Additionally, it allows inference to be multi-domain enabled and does not require user to provide domain identifiers corresponding to the input utterance. In summary, we make the following contributions: 
\begin{itemize}
    \item Build a unified neural framework to train a single semantic parser for multiple domains in the absence of ground truth parse programs. (Section~\ref{sec:proposed})
    \item We show the effectiveness of multi-policy distillation in learning a semantic parser using independent weakly supervised experts for each domain. (Section~\ref{ssec:training})
    \item We perform an extensive experimental study in  multiple domains to understand the efficacy of the proposed system against multiple baselines. We also study the effect of the availability of a small labeled corpus in the distillation setup. (Section~\ref{sec:exp})
\end{itemize}

\section{Related Work}
\begin{figure}[h]
\centering
\includegraphics[width=\columnwidth,height=5.5cm]{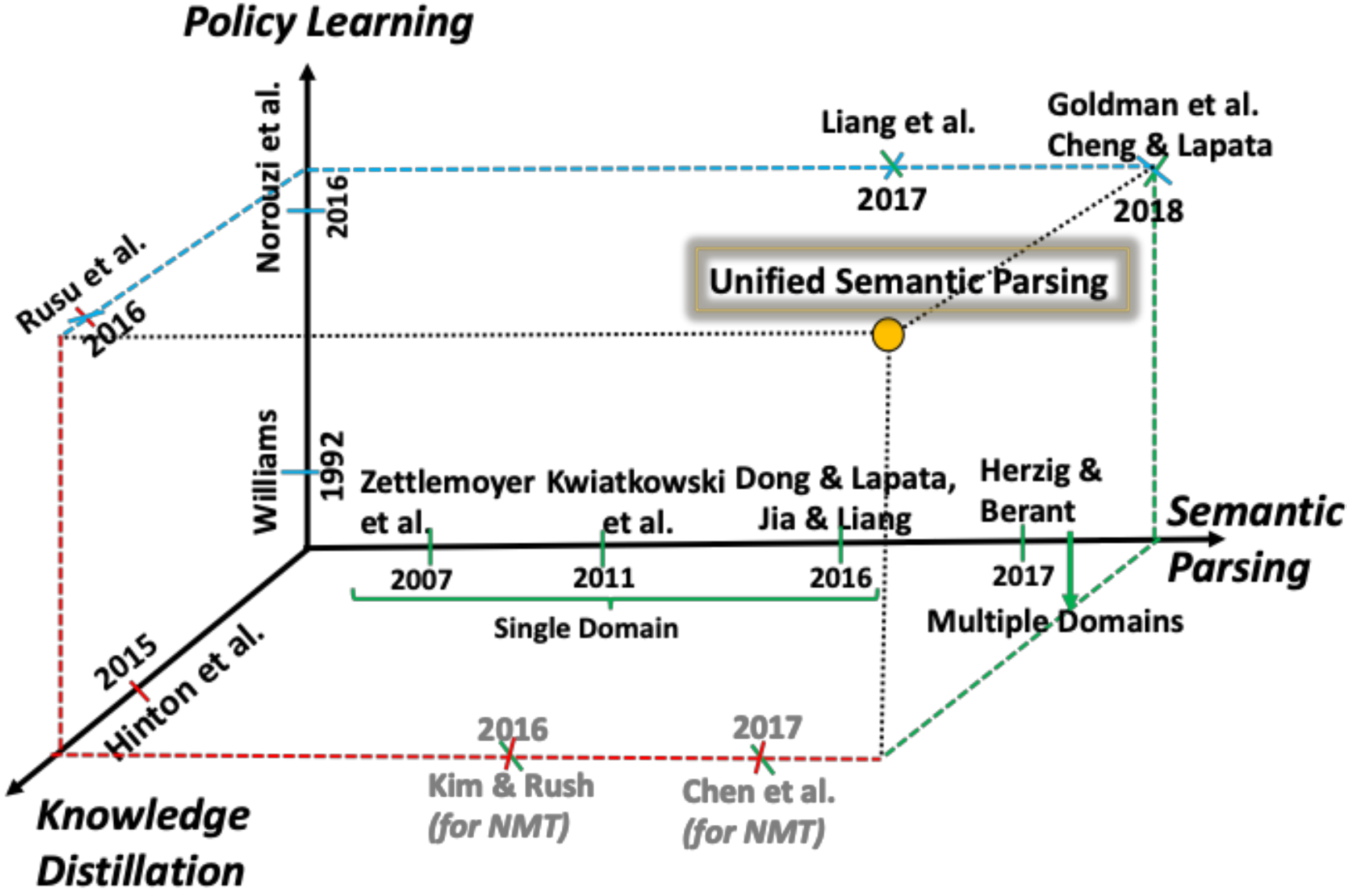}
\caption{Illustration of the proposed work in the space of key related work in the area of semantic parsing, knowledge distillation and policy learning}
\label{fig:related_work}
\end{figure}  
This work is related to three different areas: semantic parsing, policy learning and knowledge distillation.  Figure~\ref{fig:related_work} illustrates the placement of our proposed framework of unified semantic parsing in the space of the key related works done in each of these three areas. 
Semantic parsing has been an extensively studied problem,    the first study dating back to ~\newcite{Woods1973}. Much of the work has been towards exploiting annotated programs for natural language utterances to build single domain semantic parsers using various methods. ~\newcite{Zettlemoyer:2007,Kwiatkowski2011} propose to learn the  probabilistic categorical combination grammars, ~\newcite{Kate:2005} learn transformation from syntactic parse tree of natural language utterance to formal parse tree. ~\newcite{andreas2013semantic} model the task of semantic parsing as machine translation. Recently, ~\newcite{DongP16-1004} introduce the use of neural sequence-to-sequence models for the task of machine translation. Due to the cost of obtaining annotated programs, there has been an increasing interest in using weak supervision based methods ~\cite{clarke2010driving, LiangNSM, Neelakantan2016LearningAN, P18-1071, goldman2018weakly} which uses denotations, i.e. final answers obtained on executing a program on the knowledge base, for training.

The problem of semantic parsing has been primarily studied in a single domain setting employing supervised and weakly supervised techniques. However, the task of building a semantic parser in the multi-domain setting is relatively new. ~\newcite{Herzig2017NeuralSP} propose semantic parsing models using supervised learning in a multi-domain setup and is the closest to our work. However, none of the existing works inspect the problem of multi-domain semantic parsing in a weak supervision setting.

Knowledge distillation was first presented by~\newcite{Hinton} and has been popularly used for model compression of convolution neural networks in computer vision based tasks~\cite{Yu2017VisualRD,Li2017LearningFN}.  ~\newcite{kim2016seqkd,P17-1176} applied knowledge distillation on recurrent neural networks for the task of machine translation and showed improved performance with a much compressed student network. Our proposed method of policy distillation was first introduced by~\newcite{rusu-distillation-2015} and is built on the principle of knowledge distillation and applied for reinforcement learning agents. Variants of the framework for policy distillations have also been proposed~\cite{NIPS2017_7036}. To the best of our knowledge, our work is the first to apply policy distillation in a sequence-to-sequence learning task. We anticipate that the framework described in this paper can be applied to learn unified models for other tasks as well.
\section{Proposed Framework}
\label{sec:proposed}
\begin{figure*}
\centering 
    \subfigure[Domain specific expert policy $E^k$]
    {\includegraphics[width=13cm,height=3.8cm]{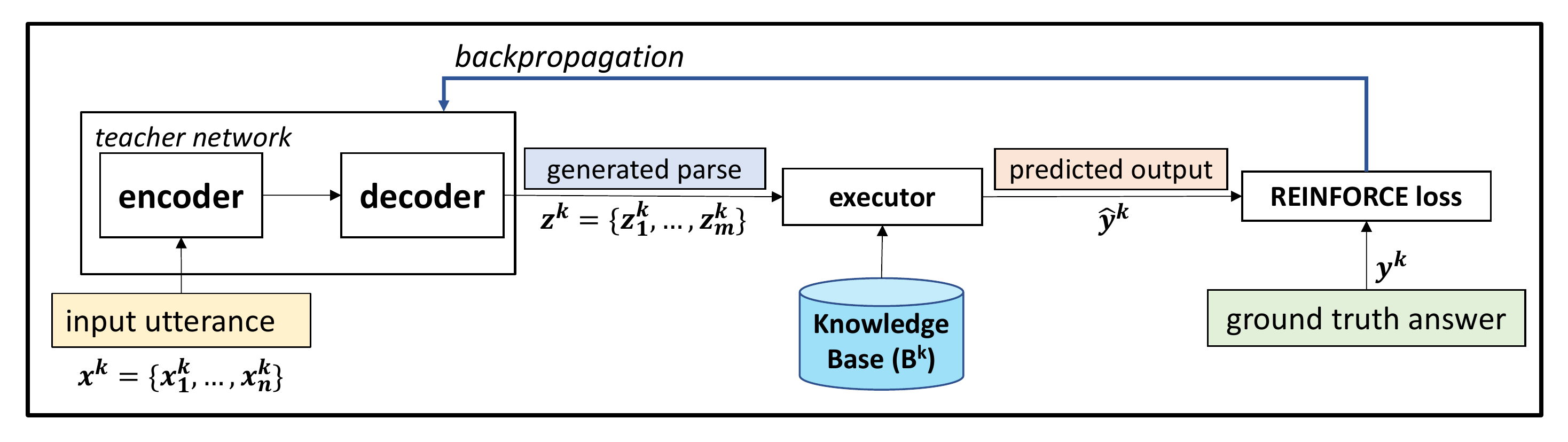}\label{fig:subfigure1_framework}}
    \subfigure[Learning a unified student $S$ by distilling domain policies from experts $E^1, \cdots ,E^K$  ]{\includegraphics[width=13cm,height=4cm]{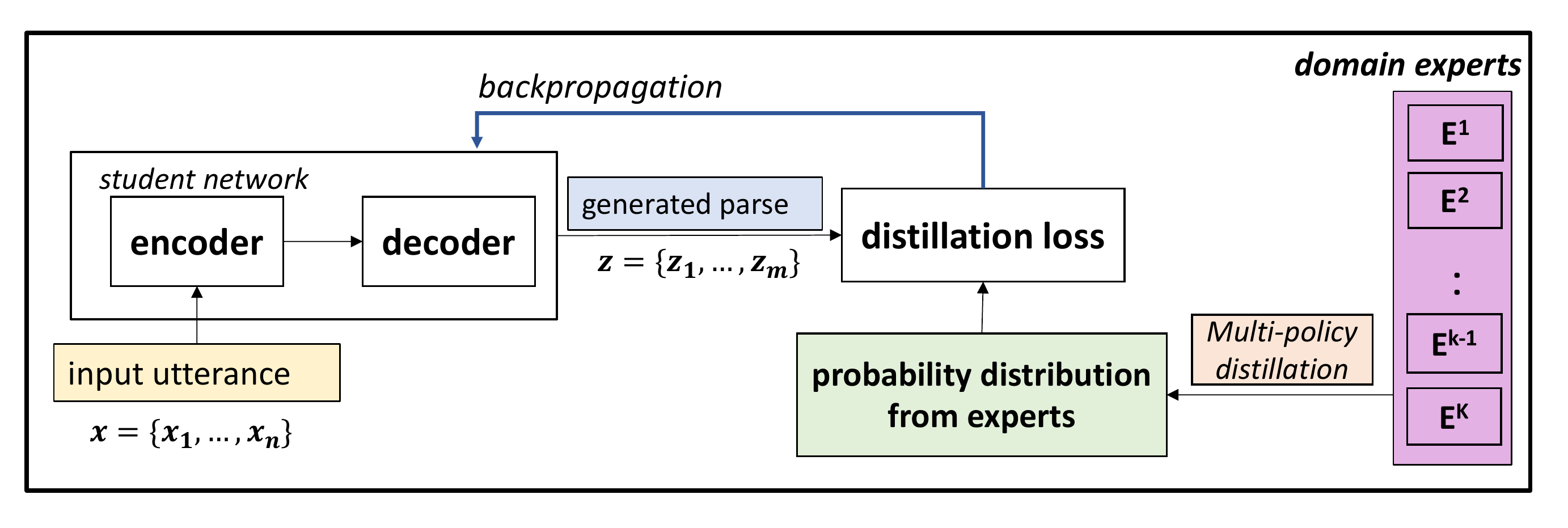}\label{fig:subfigure2_framework}}
\caption{Proposed architecture diagram of unified semantic parsing framework. Figure~\ref{fig:subfigure1_framework} demonstrates the training of the experts $E^k$ using weak supervision on the denotation corresponding to input utterance. Once we train all the domain experts $E^1, \cdots ,E^K$ for the $K$ domains, we use the probability distributions of the parse generated by these experts to train the student, thereby distilling the domain policies learnt by the teachers to the student as shown in Figure~\ref{fig:subfigure2_framework}.} 
\label{fig:architecture}
\end{figure*}

In this section, we first present a high level overview of the framework for the proposed unified semantic parsing using multi-policy distillation and then describe the models employed for each component of the framework.

 We focus on the setting of `K' domains each with an underlying knowledge-base $\mathbb{B}^1, \cdots, \mathbb{B}^K$. We have a training set of utterances $X^k$ and the corresponding final denotations $Y^k$, for each domain $k \in {1, \cdots, K}$. Unlike existing works  ~\cite{Herzig2017NeuralSP}, we do not assume availability of ground truth programs corresponding to the utterances in the training data. Our goal is to learn a unified semantic parsing model which takes as input a user utterance $x^k_i = \{x^k_{i1}, \cdots, x^k_{in}\} \in  X^k$ from any domain $k$ and produces the corresponding program $z^k_i = \{z^k_{i1},\cdots,z^k_{im}\}$ which when executed on the corresponding knowledge base $\mathbb{B}^k$ should return denotation $y_i^k \in Y^k$. In this setup, we only rely on the weak supervision from the final denotations $Y^k$ for training this model. Moreover, the domain identifier $k$ is not needed by this unified model. 

We use multi-policy distillation framework for the task of learning a unified semantic parser. Figure~\ref{fig:architecture} summarizes the proposed architecture. 
We first train parsing models (teachers) for each domain using weak supervision to learn domain-specific teacher policies. We use \textsc{REINFORCE} for training, similar to prior work on Neural Symbolic Machine \cite{LiangNSM} described briefly in Section \ref{ssec:single}. Next, we distill the learnt teacher policies to train a unified semantic parser enabled over multiple domain. (described in Section \ref{ssec:unified}).  Note that:
(1) Our teachers are trained with weak supervision from denotations instead of actual parses and hence are weaker compared to completely supervised semantic parses.
(2) State-of-the-art sequence distillation works \cite{kim2016seqkd, P17-1176} have focused on a single teacher-student setting. 

\subsection{Model}
\label{ssec:model}
In this section, we describe the architecture of semantic parsing model used for both teachers as well as the student networks. We use a standard sequence-to-sequence model~\cite{sutskever2014sequence} with attention similar to \newcite{DongP16-1004} for this task. Each parsing model (the domain specific teachers $E^1, ..., E^K$ and the unified student $S$) is composed of an $L$-layer encoder LSTM \cite{Hochreiter:1997:LSTM} for encoding the input utterances and an $L$-layer attention based decoder LSTM \cite{bahdanau+al-2014-nmt} for producing the program sequences. Note that in this section, we omit the domain id superscript $k$.

Given a user utterance $x$, the aim of the semantic parsing model is to generate output program $z$ which should ultimately result in the true denotations $y$.  This user utterance 
$x = \{x_1, ..., x_n\}$ is input to the encoder which maps each word in the input sequence to the embedding $e = \{e_1, ..., e_n\}$ and uses this embedding to update its respective hidden states $h = \{h_1, ..., h_n\}$ using $h_t = \text{LSTM}(e_t,h_{t-1};\theta_{enc})$, where $\theta_{enc}$ are the parameters of encoder LSTM. The last hidden state $h_n$ is input to the decoder's first state. The decoder updates its hidden state $s_t$ using $s_t = \text{LSTM}(c_{t-1},s_{t-1};\theta_{dec})$ where $s_{t-1}$  is the embedding of output program token $z_{t-1}$ at last step $t-1$ and $\theta_{dec}$ are the decoder LSTM parameters. 
The output program $\{z_1, ..., z_m\}$ is generated token-wise by applying softmax over the vocabulary weights derived by transforming the corresponding hidden state $s$. 

Further, we employ beam search during decoding which generates a set of parses $B$ for every utterance. At each decoding step $t$, a beam $B_t$ containing partial parses of length $t$ are maintained. The next step beam $B_{t+1}$ are the $|B|$ highest scoring expansions of programs in the beam $B_t$. 

\section{Training}
\label{ssec:training}
In this section we describe the training mechanism employed for the proposed multi-domain policy distillation framework for semantic parsing. The training process in our proposed framework has the following two components (Figure~\ref{fig:architecture}): (i) weakly supervised training for domain specific semantic parsing experts $E^1, ..., E^K$ and, (ii) distilling multiple domain policies to the unified student $S$. We next describe each of these two components.

\subsection{Domain-specific Semantic Parsing Policy}
\label{ssec:single}
As described in the previous section, an individual domain specific semantic parsing model generates
the program $z = \{z_1, ..., z_m\}$ 
which is executed on the knowledge base $\mathbb{B}$ to return the denotation $\hat y$. For brevity, we omit domain identifier $k$ and instance id $i$ in this section. In our setting, since labeled programs are not available for training, we use weak supervision from final denotations $y$ similar to \newcite{LiangNSM} for each domain expert.  As the execution of parse program is a non-differential operation on the KB,  we use \textsc{REINFORCE} \cite{williams1992simple, NIPS2016_6547} for training which maximizes the expected reward. Reward $R(x,z)$ for prediction $z$ on an input $x$ is defined as the match score between the true denotations $y$ for utterance $x$ and the denotations obtained by executing the predicted program $z$. The overall objective to maximize the expected reward is as follows
\begin{multline*}
\sum_x \mathbb{E}_{P_\theta(z|x)}[R(x,z)] \\
= \sum_x \sum_{z} {P_\theta(z|x)}R(x,z) \\
\approx \sum_x \sum_{z\in B} {P_\theta(z|x)}[R(x,z)]
\end{multline*}
where $\theta=( \theta_{enc}, \theta_{dec})$ are the policy parameters; $B$ is the output beam containing top scoring programs (described in Section \ref{ssec:model})  and $P_\theta(z|x)$ is the likelihood of parse $z$ 
\begin{equation}
\label{eq:expert}
P_\theta(z|x) = \prod_t P_\theta(z_t|x,z_{1:t-1})
\end{equation}

To reduce the variance in gradient estimation we use baseline $b(x) = \frac{1}{|B|}\sum_{z \in B} R(x, z)$ i.e. the average reward for the beam corresponding to the input instance $x$. 
See Table~\ref{table:accuracy} \textsc{WeakIndep} for the performance achieved for individual domains with this training objective. 

Note that the primary challenge with this weakly supervised training is the sparsity in reward signal given the large search space leading to only a few predictions having a non-zero reward. This can be seen in the Table~\ref{table:accuracy} \textsc{WeakCombined} when the entire set of domains is pooled into one, the numbers drop severely due to the exponential increase in the search space.

\subsection{Unified Model for multiple domains}
\label{ssec:unified}
For the unified semantic parser, we use the same sequence-to-sequence model described in Section \ref{ssec:model}.
The hyper-parameter settings vary from domain-specific models as detailed in Section \ref{ssec:impl}.
We use the multi-task policy distillation method of \newcite{rusu-distillation-2015} to train this unified parser for multiple domains. 
The individual domain experts $E^1, ..., E^K$ are trained independently as described in Section \ref{ssec:single}. This distillation framework enables transfer of knowledge from experts $E^1, ..., E^K$ to a single student model $S$ that operates as a multi-domain parser, even in the absence of any domain indicator with input utterance during the test phase. 
Each expert $E^k$ provides a transformed training dataset to the student ${D^k}=\{(x_i^k, \boldsymbol{(p^k_\theta)_i})\}_{i=1}^{|X^k|}$, where $\boldsymbol{(p^k_\theta)_i}$ is the expert's probability distribution on the entire program space w.r.t input utterance $x_i$. Concretely, given $m$ is the decoding sequence length and $\mathcal{V}$ is the vocabulary combined across domains, then $\boldsymbol{(p^k_\theta)_i} \in [0,1]^{m \times |\mathcal{V}|}$ denotes the expert $E^k$'s respective probabilities  that  output token $z_{ij}$ equals vocab token $v$, for all time steps  $j \in \{1, \dots, m \}$ and $\forall  v \in \mathcal{V}$. 
$$\boldsymbol{(p^k_\theta)_i} = \{\{ p^k_\theta(z_{ij}=v;x^k_i, z_{i\{1:j-1\}})\}_{j=1}^m \}_{v=1}^{|\mathcal{V}|} $$
The student takes the probability outputs from the experts as the ground truth and is trained in a supervised manner to minimize the cross-entropy loss $\mathcal{L}$ w.r.t to teachers' probability distribution:
\begin{multline}
\label{eq:student}
L(\theta^S; \theta^1, ..., \theta^K) = \\
- \sum^K_{k=1} \sum_{i=1}^{|X^k|} \sum_{j=1}^{|m|}\sum_{v=1}^{|\mathcal{V}|} p^k_\theta(z_{ij}=v;x^k_i, z_{i\{1:j-1\}}) \\ \log p^S_\theta(z_{ij}=v;x^k, z_{i\{1:j-1\}})
\end{multline}
where $\{\theta^k\}_{k=1}^K$ are the policy parameters of experts and $\theta^S$ are the student model parameters;   similarly $p^S_\theta(z_{ij}=v;x^k, z_{i\{1:j-1\}})$ is the probability assigned to output token $z_{ij}$ by student $S$.
This training objective enables the unified parser to learn domain-specific parsing strategies from individual domains as well as leverage structural variations across domains. 
Therefore, the combined multi-domain policy $S$ is refined and compressed during the distillation process thus rendering it to be more effective in parsing for each of the domains.

\section{Experimental Setup}
\label{sec:exp}

In this section, we provide details on the data and model used for the experimental analysis\footnote{Code and data is available at \url{https://github.com/pagrawal-ml/Unified-Semantic-Parsing}}. We further elaborate on the  baselines used.

\begin{table*}[t]
  \centering
\begin{tabular}{ lcccccc}
\hline
 \multirow{4}{*}{\textsc{DOMAIN}} & \multicolumn{3}{c}{\textsc{Original Dataset}} & \multicolumn{3}{c}{\textsc{Normalized Dataset}}\\ \cline{2-7}

& \textsc{Utterance}  & \multicolumn{2}{c}{\textsc{Program}} & \textsc{Utterance}  & \multicolumn{2}{c}{\textsc{Program}}\\ \cline{2-7}

& Vocab & Vocab & Avg.  & Vocab & Vocab & Avg. \\
&  &  & Length &  &  & Length\\ 
\hline
\textsc{Basketball} & 340 & 65 & 48.3 & 332 & 58 & 20.5\\ 
\textsc{Blocks} & 213 & 48 & 47.4 & 212 & 41	& 9.7\\
\textsc{Calendar} &	206	& 54 & 43.7 & 191 & 46 & 8.8\\
\textsc{Housing} & 302 &	58& 42.7 & 293 & 48 & 8.5\\
\textsc{Publications} & 190 & 44 & 46.2 & 187 & 38 & 8.5\\
\textsc{Recipes} &	247	& 49 & 42.6	& 241 & 40 & 7.8\\
\textsc{Restaurants}	& 315 & 62 & 41.2 & 310 & 48 & 8.2\\
\hline
\textsc{Average} & 	259	& 54.3 & 44.6 & 252.3 & 45.6 & 10.3\\
\hline
\end{tabular}
\caption{Training data statistics for original and normalized dataset. For each domain, we compare the \#unique tokens (Vocab) in input utterances and corresponding programs; and average program length.    }
  \label{tab:dataset_statistics}
\end{table*}

\subsection{Data}
We use the \textsc{Overnight} semantic parsing dataset~\cite{P15-1129} which contains multiple domains. Each domain has utterances (questions) and corresponding parses in $\lambda-$DCS form that are executable on domain specific knowledge base. Every domain is designed to focus on a specific linguistic phenomenon, for example, \textsc{calendar} on temporal knowledge, \textsc{blocks} on spatial queries. In this work, we use seven domains from the dataset as listed in Table~\ref{tab:dataset_statistics}.

We would like to highlight that we do not use the parses available in the dataset during the training of our unified semantic parser. Our weakly supervised setup uses denotations to navigate the program search space and learn the parsing policy. This search space is a function of decoder (program) length and vocabulary size. Originally, the parses have 45 tokens on an average with a combined vocabulary of 182 distinct tokens across the domains.  To reduce the decoder search space, we normalize the data to have shortened parses with an average length of 11 tokens and 147 combined vocab size.
We reduce the sequence length by using a set of template normalization functions and reduce the vocab size by masking named entities for each domain. An example of normalization function is the following: an entity utterance say of type \textit{recipe} in the query is programmed by first creating a single valued list with the entity type i.e. {\footnotesize\tt{(en.recipe)}}  and then that property is extracted : {\footnotesize\tt(call SW.getProperty ( call SW.singleton en.recipe ) ( string ! type ))} resulting in 14 tokens. We replace this complex phrasing by directly substituting  the entity type  under consideration i.e. {\footnotesize\tt (en.recipe)}  (1 token). Next, we show an example for a complete utterance:  \textit{what recipes posting date is at least the same as rice pudding}. Its original parse is: 
\begin{Verbatim}[fontsize=\small]
(call SW.listValue (call SW.filter 
(call SW.getProperty (call SW.singleton 
 en.recipe) (string ! type)) (call 
SW.ensureNumericProperty (string 
posting_date)) (string >=) 
(call SW.ensureNumericEntity (call 
SW.getProperty en.recipe.rice_pudding 
(string posting_date))))).
\end{Verbatim}
Our normalized query is \textit{ what recipes posting date is at least the same as e0}, where entity \textit{rice pudding} is substituted by entity identifier \textit{e0}. The normalized parse is as follows:	
\begin{Verbatim}[fontsize=\small]
SW.filter en.recipe 
 SW.ensureNumericProperty 
    posting_date >= 
        (SW.ensureNumericEntity 
         SW.getProperty e0 posting_date)
\end{Verbatim}

It is important to note that this normalization function is reversible. During the test phase, we apply the reverse function to convert the normalized parses to original forms for computing the denotations.  Table \ref{tab:dataset_statistics} shows the domain wise statistics of original and normalized data. It is important to note that this script is applicable for template reduction for any $\lambda-$DCS form.

We report hard denotation accuracy i.e. the proportion of questions for which the top prediction and ground truth programs yield the matching answer sets as the evaluation metric. For computing the rewards during training, we use soft denotation accuracy i.e. F1 score between predicted and ground truth answer sets.

Table~\ref{table:accuracy} shows the accuracy with strongly supervised training (\textsc{Supervised}). The average denotation accuracy (with beam width 1) of 70.6\% which is comparable to state-of-the-art \cite{P16-1002} denotation accuracy of 75.6\% (with beam width 5). This additionally suggests that data normalization process does not alter the task complexity. 


\subsection{Baselines}
\label{ssec:baselines}
In the absence of any work on multi-domain parser trained without ground truth programs, we compare the performance of the proposed unified framework against the following baselines:
\begin{enumerate}
    \item \textbf{Independent Domain Experts} (\textsc{Weak-Independent}): These are the set of weakly supervised semantic parsers, trained independently for each domain using \textsc{REINFORCE} algorithm as described in Section~\ref{ssec:single}. Note that these are the teachers in our multi-policy distillation framework.
    \item \textbf{Combined Weakly Supervised Semantic Parser} (\textsc{Weak-Combined})): As per the recommendation in \newcite{Herzig2017NeuralSP}, we pool all the domains datasets into one and train a single semantic parser with weak supervision. 
    \item \textbf{Independent Policy Distillation} (\textsc{Distill-Independent}): We also experiment with independent policy distillation for each domain. The setup is similar to the one described in Section~\ref{ssec:unified} used to learn $K$ student parsing models, one for each individual domain. Each student model uses the respective expert model as the only teacher. 
\end{enumerate}
Following the above naming convention, we term our proposed framework as \textsc{Distill-Combined}.
For the sake of completeness, we also compute the skyline \textsc{Supervised } i.e. the sequence-to-sequence model described in Section~\ref{ssec:model} trained with ground truth parses. 

\subsection{Model Setting}
\label{ssec:impl}
We use the original train-test split provided in the dataset. We further split the training set of each domain into training (80\%) and validation (20\%) sets. We tune each hyperparameter by choosing the parameter from a range of values and choose the configuration with highest validation accuracy for each model. For each experiment we select from: beam width = \{1, 5, 10, 20\}, number of layers = \{1,2,3,4\}, rnn size for both encoder \& decoder = \{100, 200, 300\}.
For faster compute, we use the string match accuracy as the proxy to denotation reward. 
In our experiments, we found that combined model performs better with the number of layers set to 2 and RNN size set to 300 while individual models' accuracies did not increase with an increase in model capacity. This is intuitive as the combined model requires more capacity to learn multiple domains. Encoder and decoder maximum sequence lengths were set to 50 and 35 respectively. For all the models, RMSprop optimizer \cite{hinton2012neural} was used with learning rate set to 0.001. 

\section{Results and Discussion}
\label{ssec:result}
Table \ref{table:accuracy} summarizes our main experimental results. It shows that our proposed framework \textsc{Distill-Combined} clearly outperforms the three baselines \textsc{Weak-Independent, Weak-Combined, Distill-Independent} described in Section~\ref{ssec:baselines}  

\begin{table*}[t]
\centering
\begin{tabular}{lcccc|c} 
\toprule
\multirow{2}{*}{\textsc{DOMAIN}} & \textsc{Weak-} & \textsc{Weak-}  & \textsc{Distill-} & \textsc{Distill-} &  \\
  & \textsc{Independent}  & \textsc{Combined} & \textsc{Independent} & \textsc{Combined} & \textsc{Supervised}  \\
 \midrule
\textsc{Basketball} & 33.8 & 0.5 & 33.8 & \textbf{36.3} & 81.0 \\ 
\textsc{Blocks} & 27.6 & 0.8 & 36.8 & \textbf{37.1} & 52.8\\
\textsc{Calendar} & \textbf{25.0} & 0.6 & 12.5 & 17.3 & 72.0\\
\textsc{Housing} & 33.3 & 2.1 & 42.3 & \textbf{49.2}  & 66.1\\
\textsc{Publications} & 42.2 & 6.2 & 45.9 & \textbf{48.4} & 68.3\\  
\textsc{Recipes} & 45.8 & 2.3 & 61.5  & \textbf{66.2} & 80.5\\
\textsc{Restaurants} & 41.3 & 2.1 & 40.9 & \textbf{45.2} & 73.5\\
\hline
\textsc{Average} & 35.5 & 2.1 & 39.1 & \textbf{42.8} & 70.6\\
\bottomrule
\end{tabular}
\caption{Test denotation accuracy for each domain comparing our proposed method \textsc{DistillCombined} with the three baselines. We also report the skyline \textsc{Supervised}.}
\label{table:accuracy}
\end{table*}

\textbf{Effect of Policy Distillation:}   \textsc{Distill-Independent} are individual domain models trained through distillation of individual weakly supervised domain experts policies \textsc{Weak-Independent}. We observe that policy distillation of individual expert policies result in an  average percentage increase of $\sim10\%$ in accuracy with a maximum of $\sim33\%$ increase in case of \textsc{blocks} domains, which shows the effectiveness of the distillation method employed in our framework. Note that for \textsc{calendar} domain, \textsc{Weak-Independent} is unable to learn the parsing policy probably due to the complexity of temporal utterances. Therefore, further distillation on the inaccurate policy leads to drop in performance. More systematic analysis on the failure cases is an interesting future direction.  

\textbf{Performance of Unified Semantic Parsing framework:} The results show  the proposed unified semantic parser using multi-policy distillation (\textsc{Distill-Combined}) (as described in section \ref{sec:proposed}) on an average has the highest performance in predicting programs under weak supervision setup.  \textsc{Distill-Combined} approach leads to an increased performance by $\boldmath\sim20\%$   on an average in comparison to individual domain specific teachers (\textsc{Weak-Independent}). We note maximum increase in the case of \textsc{Housing} domain with $\boldmath\sim47\%$ increase in the denotation accuracy. 

\textbf{Effectiveness of Multi-Policy Distillation:} Finally, we evaluate the effectiveness of the overall multi-policy distillation process in comparison to training a combined model with data merged from all the domains (\textsc{Weak-Combined}) in the weak supervision setup. We observe that due to weak signal strength and enlarged search space from multiple domains, \textsc{Weak-Combined} model performs poorly across domains. Thus, further reinforcing the need for the distillation process. 
As discussed earlier, the \textsc{Supervised} model is trained using strong supervision from ground-truth parses and hence is not considered as a comparable baseline, rather a skyline, for our proposed model

\begin{figure}
\centering
\includegraphics[width=\columnwidth]{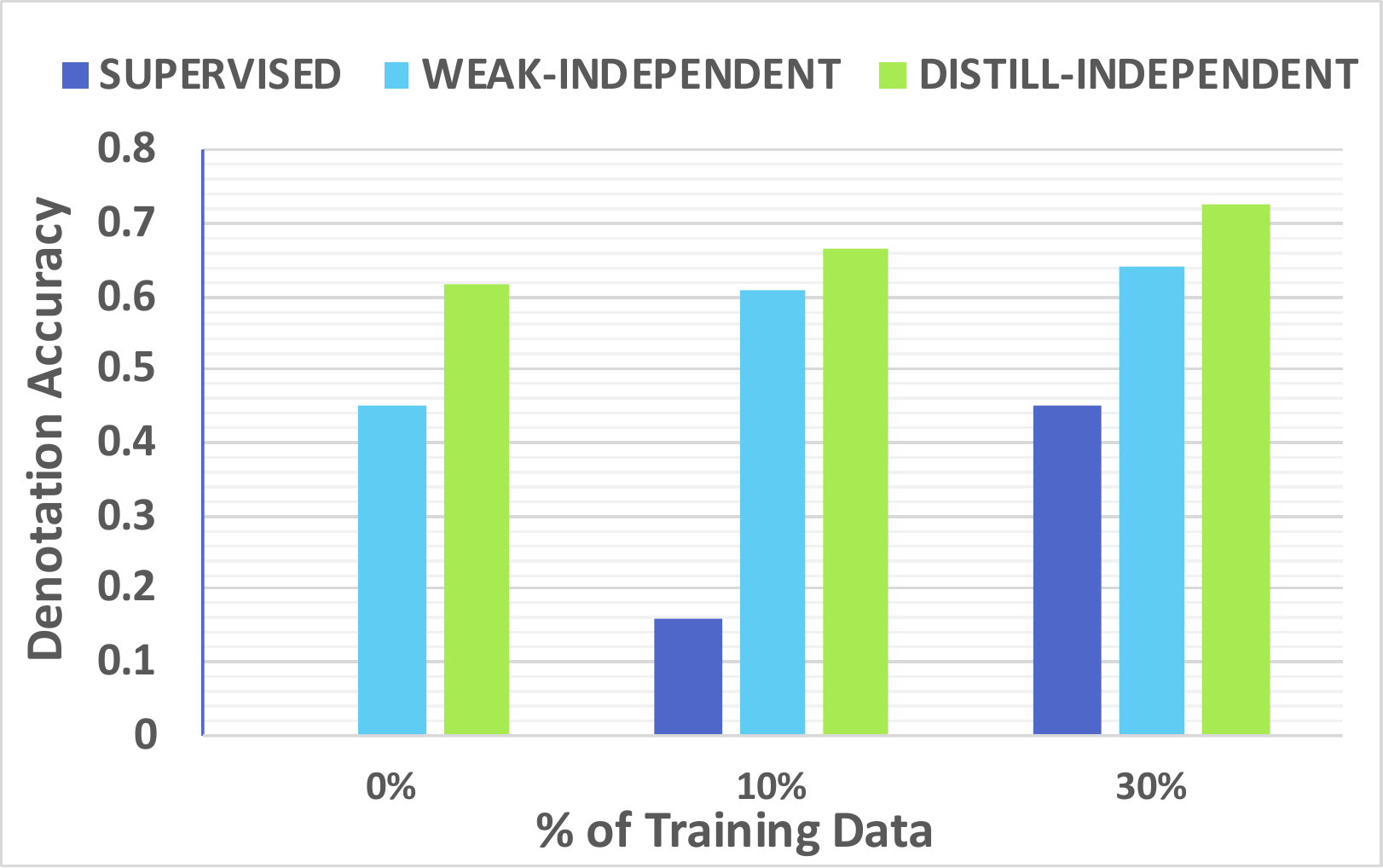}
\caption{Effect of the fraction of training data on different models}
\label{fig:chart_pretrain}
\end{figure} 

\subsection{Effect of Small Parallel Corpus} We show that our model can greatly benefit from the availability of a limited amount of parallel data where semantic parses are available. Figure \ref{fig:chart_pretrain} plots the performance of \textsc{Weak-Independent} and \textsc{Distill-Independent} models for \textsc{recipes} domain when initialized with a pre-trained  \textsc{Supervised} model trained on 10\% and 30\% of parallel training data. As it can be seen, adding 10\% parallel data brings an improvement of about 5 points, while increasing the parallel corpus size to only 30\% we observe an improvement of about 11 points. The observed huge boost in performance is motivating given the availability of small amount of parallel corpus in most real world scenarios.
%

\section{Conclusions and Future Work}
In this work, we addressed the challenge of training a semantic parser for multiple domains without strong supervision i.e. in the absence of ground truth programs corresponding to input utterances. 
We propose a novel unified neural framework using multi-policy distillation mechanism with two stages of training through weak supervision from denotations i.e. final answers corresponding to utterances. The resultant multi-domain semantic parser is compact and more precise as demonstrated on the \textsc{Overnight} dataset.
We believe that this proposed framework has wide applicability to any sequence-to-sequence model. 

We show that a small parallel corpus with annotated programs boosts the performance. We plan to explore if further fine-tuning using  denotations based training on the distilled model can lead to improvements in the unified parser.  We also plan to investigate the possibility of augmenting the parallel corpus by bootstrapping from shared templates across domains. This would further make it feasible to perform transfer learning on a new domain. An interesting direction would be to enable domain experts to identify and actively request for program annotations given the knowledge shared by other domains. We would also like to explore if guiding the decoder through syntactical and domain-specific constraints helps in reducing the search space for the weakly supervised unified parser.  

\section*{Acknowledgement}
We thank Ghulam Ahmed Ansari and Miguel Ballesteros, our colleagues at IBM for discussions and suggestions which helped in shaping this paper.

\bibliography{acl2019}
\bibliographystyle{acl_natbib}

\end{document}